\documentclass[10pt]{article} 
\usepackage[preprint]{tmlr}
\usepackage{graphicx} 

\usepackage{amsmath,amsfonts,bm}

\def\eqref#1{equation~\ref{#1}}

\def\1{\bm{1}}

\DeclareMathAlphabet{\mathsfit}{\encodingdefault}{\sfdefault}{m}{sl}
\SetMathAlphabet{\mathsfit}{bold}{\encodingdefault}{\sfdefault}{bx}{n}

\usepackage{algorithm}
\usepackage{algpseudocode}
\usepackage{placeins}
\usepackage{float}
\usepackage{hyperref}
\usepackage{url}

\title{Enabling Robust In-Context Memory and Rapid Task Adaptation in Transformers with Hebbian and Gradient-Based Plasticity}

\author{\name Siddharth Chaudhary \email chaudhs@stpaulsschool.org.uk \\
      \addr St Paul's School, London
      }

\def\month{MM}  
\def\year{YYYY} 
\def\openreview{\url{https://openreview.net/forum?id=XXXX}} 

\begin{document}

\maketitle

\begin{abstract}
Large language models display in-context learning as an emergent effect of scale, but they rely on static weights during inference. In contrast, biological systems continually adapt via synaptic plasticity. We investigate whether explicit, biologically inspired plasticity can endow Transformers with faster in-sequence adaptation. To this end, we augment decoder-only Transformers with fast-weight modules updated either by (i) a neuromodulated Hebbian rule or (ii) the gradient-based plasticity mechanism of Duan et~al.\ (2023). Across copying, regression, and few-shot classification tasks (CIFAR-FS, Omniglot), Hebbian plasticity consistently achieves lower loss and stronger few-shot generalization, while gradient-based updates perform best on long-horizon credit assignment. When associations are short and linearly separable, static weights suffice, defining a clear boundary condition for when plasticity helps. Analysis of learned modulatory signals reveals that gradient-based rules maintain large, persistent updates, whereas Hebbian plasticity is sharply gated around salient events. Together, these results show that explicit plasticity complements attention by enabling rapid, task-specific adaptation, and clarify when different plasticity mechanisms are most effective.

\end{abstract}

\section{Introduction}

Transformers exhibit striking in-context learning (ICL) capabilities: given a few input–output examples, they can perform novel tasks without weight updates. Yet this ability is largely an emergent consequence of scale and training diversity rather than an explicit mechanism for adaptation \citep{brown2020language, dai2023can}. As a result, the model's capacity to incorporate new evidence is limited to transient activations within self-attention, with no dedicated process for storing or consolidating information during inference.

Biological neural systems, in contrast, continuously modify synaptic strengths in response to experience \citep{magee2020synaptic}. Such \emph{synaptic plasticity} enables rapid, context-dependent learning and flexible generalization \citep{bittner2017behavioral}. Inspired by this principle, we ask: \emph{Can explicit plasticity mechanisms improve a Transformer's ability to adapt within a sequence?}

To answer this, we augment decoder-only Transformers with \emph{fast-weight components} that are updated online at each generation step. These updates follow one of two biologically motivated rules: (i) a \textbf{neuromodulated Hebbian rule}, which adjusts weights locally based on correlated pre- and post-synaptic activity, and (ii) a \textbf{gradient-based plasticity rule} \citep{duan2023hebbian}, which performs local gradient descent on an internally generated target signal. This formulation allows Transformers to adapt during inference, not just across training steps.

We evaluate these plastic Transformers on the same family of memory and few-shot learning benchmarks previously used for plastic RNNs, enabling a direct comparison across architectural priors. Our analysis reveals complementary regimes of effectiveness: gradient-based plasticity excels on tasks requiring long-range credit assignment, while Hebbian updates dominate in structured, sparse-supervision settings such as few-shot classification and regression.

Our main contributions are:
\begin{enumerate}
    \item A general framework for incorporating step-wise synaptic plasticity into autoregressive Transformers.
    \item A systematic comparison of Hebbian and gradient-based plasticity across copying, regression, and meta-learning benchmarks.
    \item Mechanistic insights into how explicit plasticity interacts with self-attention dynamics, clarifying when and why it benefits in-context learning.
\end{enumerate}

Together, these results bridge the gap between emergent and mechanistic adaptation, showing that explicit synaptic plasticity can endow Transformers with more efficient, interpretable forms of in-sequence learning.

\section{Related Work}

Our research is situated at the intersection of three active fields: meta-learning, the study of dynamic parameters in artificial networks, and the analysis of in-context learning in Transformers.

\subsection{Meta-Learning for Rapid Adaptation}
Our work falls under the "learning to learn" or meta-learning paradigm \cite{schmidhuber1997shifting, bengio1992optimization}, which aims to build models that can adapt to new tasks from limited experience. This field is characterized by a "bi-level" optimization process: an "inner loop" where the model adapts to a specific task, and an "outer loop" where the meta-learner is updated to improve this adaptation process. Several distinct approaches have emerged:

\begin{itemize}
    \item \textbf{Gradient-based Meta-Learning:} This line of work, exemplified by Model-Agnostic Meta-Learning (MAML) \cite{finn2017model}, meta-learns a parameter initialization $\theta$ that can be effectively fine-tuned to a new task with only a few gradient steps. This approach is powerful but typically assumes that the inner-loop task provides an explicit, differentiable loss function.

    \item \textbf{Metric-based Meta-Learning:} Methods like Prototypical Networks \cite{snell2017prototypical} learn a deep embedding space where new examples can be classified based on their proximity to "prototypes" (e.g., class means) derived from the support set. While effective, these methods are often specialized for few-shot classification.

    \item \textbf{Memory-Augmented Networks (MANNs):} These models, such as those proposed by Santoro et al. \cite{santoro2016meta}, utilize an external memory (e.g., a Neural Turing Machine's memory bank) to store task-specific information, which the controller network can read from and write to.

    \item \textbf{Meta-Learning a Learning Rule:} Our work, following Duan et al. \cite{duan2023hebbian}, belongs to a category that meta-learns the \textit{learning rule itself}. Instead of learning an initialization (like MAML) or an embedding (like ProtoNets), the outer loop meta-trains the parameters of a dynamic update rule (e.g., the connection-specific learning rates $\alpha$).
\end{itemize}

The primary advantage of this last approach is its generality and biological plausibility. The plasticity rules are task-agnostic and operate in an unsupervised fashion, driven only by network activity or a self-generated target. This bypasses the need for an explicit, human-defined loss function in the inner loop, allowing the model to adapt to arbitrary sequential experiences.

\subsection{Dynamic Parameters and Synaptic Plasticity}
The core mechanism we explore is the use of "fast weights"—parameters that change on a rapid timescale (i.e., step-by-step) within a single forward pass, as opposed to the "slow weights" updated by the outer loop's gradient descent. This concept has two main branches in the literature:

\begin{itemize}
    \item \textbf{Biologically-Inspired Plasticity:} This approach draws direct inspiration from neuroscience. Hebb's rule ("cells that fire together, wire together") is a classic model of such plasticity. In ANNs, this has been adapted to create associative memories \cite{limbacher2020harnessing, schlag2021b} and to meta-learn policies in simple reinforcement learning tasks \cite{najarro2020meta}. The most direct precursors to our work are Miconi et al. \cite{miconi2018differentiable, miconi2019backpropamine} and the original paper by Duan et al. \cite{duan2023hebbian}, which applied differentiable, generalized Hebbian rules to RNNs to solve memory and meta-learning problems.

    \item \textbf{Hypernetworks and Fast Weight Programmers:} An alternative, engineering-driven approach is to use a secondary network, or "hypernetwork," to generate the weights of a primary network \cite{ha2017hypernetworks}. This concept has been adapted for Transformers, notably in "Fast Weight Programmers" \cite{schlag2021linear, irie2021going}, where a recurrent network or attention mechanism generates the fast weights used by another component.
\end{itemize}

Our work differs from this latter branch by not using a separate network to \textit{generate} weights. Instead, we implement specific, local (Hebbian) and global (gradient-based) \textit{update rules} that modify the weights based on the network's own activity, as proposed in \cite{duan2023hebbian}.

\paragraph{Relation to test-time training (TTT).} Our gradient-based plasticity is related in spirit to test-time adaptation methods that update parameters using an internally defined objective during inference (often called test-time training or adaptation). However, there are key differences: (i) our inner objective $L(t)$ is computed \emph{per step} from variables already emitted by the model and targets only the fast plastic state; (ii) the outer objective remains the task loss and trains the \emph{rules} and slow parameters that make such step-wise adaptation useful; and (iii) our updates are neuromodulated and gated, not unconditional. Recent work on TTT \citep{dalal2025tttvideo, behrouz2024titans} typically optimizes a self-supervised auxiliary loss on the \emph{same} slow parameters at test time, whereas here only fast weights are updated online while slow weights remain fixed.

\subsection{Explicit vs. Implicit In-Context Learning in Transformers}
A significant challenge and inspiration for this work is the remarkable in-context learning (ICL) ability of large-scale Transformers \cite{brown2020language}. These models appear to learn new tasks at inference time from examples in their context, all \textit{without} any explicit fast weights or gradient updates. This has led to two diverging views on the mechanism of ICL:

\begin{itemize}
    \item \textbf{The "Implicit" View (ICL as Optimization):} A prominent theoretical line analyzes the Transformer's forward pass as an implicit optimization algorithm. From this perspective, the self-attention mechanism learns to implement a learning algorithm (e.g., a form of gradient descent \cite{garg2022transformers} or Bayesian inference \cite{xie2022explanation}) on the context. The model's parameters (slow weights) are trained to encode an algorithm, such as ridge regression, which is then \textit{executed} in the forward pass using the context examples \cite{akyurek2023learning}. In this view, "learning" is an ephemeral process that exists only in the activation patterns and attention matrices.

    \item \textbf{The "Explicit" View (ICL as Plasticity):} This is the "dense-versus-sparse" taxonomy we explore. The implicit view describes a "sparse" update, where learning is encoded in activations. We investigate the "dense" alternative: modifying the network's weights themselves. Our work tests whether augmenting a Transformer with the \textit{explicit} plasticity mechanisms from Duan et al. \cite{duan2023hebbian} provides a more direct, robust, or efficient mechanism for ICL. Note that standard Transformers use only the implicit mechanism; the explicit variant refers to our plastic-augmented model.
\end{itemize}

Our contribution is to bridge this gap. We directly compare a standard Transformer (our baseline, which performs "implicit" ICL) against "Plastic Transformers" of the same parameter count that are equipped with explicit, persistent weight modifications $w(t)$. This allows us to test whether these biologically-inspired rules offer a complementary or superior method for in-context adaptation compared to the emergent "implicit" optimization of standard Transformers.

\section{Method}
\label{sec:method}

\subsection{Overview}

We endow a standard decoder-only Transformer with \emph{fast weights} that adapt during inference via biologically inspired update rules. Each plastic layer maintains a short-term weight memory that evolves as the sequence unfolds, enabling the model to incorporate new information without gradient updates on the static parameters. The outer (meta) loop optimizes the static parameters to support these in-sequence adaptations.

\subsection{Model Framework}

We modify the position-wise feed-forward networks (FFNs) to include a plastic component. For any plastic layer $l$, the effective weights at token position $t$ are
\begin{equation}
    W_l(t) = \tilde{W}_l + w_l(t),
\end{equation}
where $\tilde{W}_l$ are static, meta-trained parameters and $w_l(t)$ are fast weights initialized to zero at the start of each new sequence. The static parameters---including self-attention matrices and layer-normalization weights---are optimized by outer-loop backpropagation, while the plastic components evolve according to a local update rule during the forward pass.

\paragraph{Meta-training procedure.}
Plastic Transformers are trained using a two-timescale optimization loop (Algorithm~\ref{alg:training}). Within each sequence (the \emph{inner loop}), the fast weights are updated at every step using either a Hebbian or gradient-based plasticity rule. Across sequences, an \emph{outer loop} optimizes the static parameters $\tilde{W}_l$ and connection-specific learning rates $\alpha_l$ to minimize the overall meta-loss. This setup parallels learned-optimizer frameworks and the plastic-RNN training of \citet{duan2023hebbian}.

\begin{algorithm}[h!]
\caption{Outer--Inner Meta-Training for Plastic Transformers}
\label{alg:training}
\begin{algorithmic}[1]
    \While{not converged}
        \State Sample batch of sequences $\mathcal{T}_i \sim \mathcal{T}$
        \For{each sequence $\mathcal{T}_i$}
            \State Initialize fast weights $w_l(0)\!\leftarrow\!0$ for all plastic layers
            \For{time step $t = 1..T$}
                \State Compute output $o_t = f(x_t; \tilde{W} + w(t))$
                \State Compute modulation signal $\eta(t)$
                \State Update $w_l(t) \!\leftarrow\! (1 - \eta(t))w_l(t\!-\!1) + \eta(t)\alpha_l \!\circ\! \Delta w_l(t)$
            \EndFor
        \EndFor
        \State Update static parameters $\tilde{W}_l, \alpha_l$ via gradient descent on accumulated meta-loss
    \EndWhile
\end{algorithmic}
\end{algorithm}

At each step, the model outputs both the token prediction $y_t$ and auxiliary variables $(\tilde{\eta}_t, \overline{y}_t)$ that control the internal modulation and gradient-based updates. Here, $\tilde{\eta}_t$ is a scalar neuromodulation \emph{logit} (pre-sigmoid) and $\overline{y}_t\in \mathbb{R}^{d_{\text{aux}}}$ is an auxiliary vector produced by a small head alongside $y_t$. The additional state for plasticity consists of the fast weights $w_l(t)$ (and biases) matching the shape of the corresponding static weights, i.e., $O(d_{\text{model}}^2)$ parameters per plastic FFN matrix. This adds a small, task- and configuration-dependent overhead in runtime and memory; the code paths make this explicit, and we quantify sizes below when discussing each rule.

\subsection{Hebbian Plasticity}

For a plastic FFN layer $l$, let $p_l(t)$ denote the pre-synaptic activations (layer input) and $q_l(t)$ the post-synaptic activations (layer output). The Hebbian update computes an activity-dependent weight increment:
\begin{equation}
    w_{l}(t+1) = (1-\eta(t))w_{l}(t) + \eta(t)\,\alpha_{l} \circ \big(p_{l}(t)q_{l}^{\top}(t)\big),
\end{equation}
where $\alpha_l$ are learnable per-connection learning rates and $\eta(t)$ is a global, learned neuromodulation factor controlling when plasticity is active. The operator $\circ$ denotes elementwise (Hadamard) product and $\text{Vec}(\cdot)$ flattens a matrix. Following \citet{duan2023hebbian}, we compute
\begin{align}
    \eta(t) &= \eta_{0}\,\sigma(\tilde{\eta}_{t}) \times 
    \min\!\left(1,\frac{\text{max\_norm}}{\lVert\delta_t\rVert_2}\right), \label{eq:eta}\\
    \delta_t &= \text{Concat}\!\big(\text{Vec}(p_l(t)q_l^{\top}(t))\mid l\in S\big),
\end{align}
where $\sigma$ is the sigmoid, $\eta_0$ is a scalar hyperparameter, and $S$ is the set of plastic layers. This mechanism allows the network to form and gate short-term associations based on contextual relevance.

\subsection{Gradient-Based Plasticity}

The second rule replaces the Hebbian correlation with an internally generated gradient signal. The model constructs an auxiliary loss $L(t)$ from its own outputs:
\begin{equation}
    L(t) = \frac{1}{d_o}\left\lVert W_{\text{out}}^{\top}\,\text{Concat}(y_t,\overline{y}_t,\tilde{\eta}_t)\right\rVert_2^2,
\end{equation}
where $d_o$ is the output dimensionality. The fast weights and biases are then updated by taking a neuromodulated gradient step on this internal loss:
\begin{align}
    w_l(t\!+\!1) &= (1-\eta(t))w_l(t) + \eta(t)\,\alpha_l \circ \frac{\partial L(t)}{\partial w_l(t)}, \\
    b_l(t\!+\!1) &= (1-\eta(t))b_l(t) + \eta(t)\,\beta_l \circ \frac{\partial L(t)}{\partial b_l(t)}.
\end{align}
Here $\beta_l$ are learnable bias-specific rates, and $\eta(t)$ is computed as in Eq.~\ref{eq:eta} with $\delta_t$ defined as the concatenation of all internal gradients. $W_{\text{out}}\in\mathbb{R}^{d_o\times d_o}$ is a learned square projection mapping the concatenated vector $\text{Concat}(y_t,\overline{y}_t,\tilde{\eta}_t)\in\mathbb{R}^{d_o}$ back to $\mathbb{R}^{d_o}$; here $d_o = d_{\text{class}} + d_{\text{aux}} + 1$. This rule gives the model an explicit mechanism for credit assignment within the current sequence, complementing the associative memory provided by the Hebbian rule.

\paragraph{Computational complexity and overhead.} Memory overhead comes from storing fast weights $w_l(t)$ (and biases) with the same shape as their static counterparts. For a plastic FFN with two linear maps of shapes $(d_{\text{ff}}, d_{\text{model}})$ and $(d_{\text{model}}, d_{\text{ff}})$, the additional state per block is $O(d_{\text{ff}}\,d_{\text{model}})$ parameters per matrix (plus optional biases). Hebbian updates add outer-products and elementwise operations per step (roughly $O(d_{\text{ff}}\,d_{\text{model}})$ per plastic matrix). The gradient-based rule additionally computes an internal projection on $d_o$-dimensional vectors (cost $O(d_o^2)$ per step, typically small) and an autograd pass to obtain $\partial L/\partial w$, whose cost is comparable to a lightweight backward through the plastic FFN. In all cases, slow (static) parameters are unchanged during inference; only the fast state is updated online.

\section{Experiments}

We evaluate plastic Transformers on the same suite of tasks introduced by \citet{duan2023hebbian}, testing two hypotheses:  
(1) explicit plasticity improves short-term memory, and  
(2) it enables rapid, in-sequence learning.  
All data generation, losses, and evaluation protocols follow the original setups, with model sizes matched in parameter count to ensure comparability.

\subsection{Experimental Protocol}

Each configuration is run with $S{=}3$ independent seeds. Reported values are mean $\pm$ standard deviation across seeds.

Evaluation isolates plasticity: every experiment includes a non-plastic baseline (\texttt{rule=none}) trained with identical optimizer, architecture, and data protocol.

Inner vs. outer updates. Within each sequence/episode (inner loop), the plastic state is initialized to zero and updated \emph{online at every step} via either the Hebbian increment or the gradient-based step on $L(t)$. The static parameters (including $\tilde{W}$ and the learning-rate tensors $\alpha,\beta$) are updated only by the outer loop, \emph{once per sequence}, using the task loss accumulated across that sequence. At test time, no outer-loop updates are applied: models adapt only through their in-sequence plasticity.

Data generation and dataset sizes. For synthetic tasks (copying, cue–reward, regression), episodes are generated \emph{on the fly} in \texttt{\_\_getitem\_\_}, so there is no fixed set to memorize; \texttt{dataset\_size} controls the number of fresh sequences drawn per epoch. For the reported runs we use: 50 sequences (copying; delay 20), 100 cue–reward trials, 150 regression episodes, and 80/100 classification episodes per epoch (CIFAR-FS/Omniglot). Each sequence is seen once per epoch for $E$ epochs (see schedule below), with state reset between sequences. This balances compute and variance while keeping protocols consistent across rules.

Inner update counts per task. The number of inner (plastic) updates per episode equals the sequence length $T$:
\begin{itemize}
    \item Copying: $T = 2\,\text{seq\_length} + \text{delay} + 1$ (presentation, delay, delimiter, recall).
    \item Cue–reward: $T = 2\,\text{num\_pairs}$ (present then query for each cue).
    \item Few-shot regression: $T = K_{\text{support}} + K_{\text{query}}$.
    \item Few-shot classification: $T = N\!\times\!K$ (support) $+\;N\!\times\!Q$ (queries) for $N$-way, $K$-shot, $Q$-query.
\end{itemize}

Per-task training schedule (as used in released artifacts). Copying: 50 episodes/epoch for $E{=}2$ epochs (total $100$ episodes/run). Cue–reward: 100 episodes/epoch, $E{=}4$. Regression: 150 episodes/epoch, $E{=}5$. CIFAR-FS: 80 episodes/epoch, $E{=}5$. Omniglot: 100 episodes/epoch, $E{=}5$. Outer-loop updates occur once per episode; inner updates occur $T$ times per episode as above.

Per-step neuromodulation $\eta(t)$ and the Frobenius norm of each plastic matrix are logged for all runs.  
We consider an improvement \emph{reliable} when the mean difference exceeds the pooled standard deviation across seeds and ordering is consistent for all runs.

\subsection{Copying Task}
\textbf{Task.} Reproduce a random sequence (length $n=5$) after a 20-step delay ($m=20$).  
\textbf{Metric.} Validation loss and recall accuracy (higher is better).

\begin{table}[t]
\caption{Copying task ($n=5$, $m=20$). Mean $\pm$ std over three seeds.}
\label{tab:copying}
\begin{center}
\begin{tabular}{lcc}
\multicolumn{1}{c}{\bf RULE} & \multicolumn{1}{c}{\bf LOSS} & \multicolumn{1}{c}{\bf RECALL ACCURACY}
\\ \hline \\
Gradient & 0.352 $\pm$ 0.021 & {\bf 0.745 $\pm$ 0.011} \\
Hebbian & {\bf 0.345 $\pm$ 0.009} & 0.727 $\pm$ 0.007 \\
None & 0.415 $\pm$ 0.105 & 0.721 $\pm$ 0.025 \\
\end{tabular}
\end{center}
\end{table}

\noindent
\textbf{Interpretation.}
Both plasticity rules outperform the non-plastic baseline. Gradient-based plasticity achieves the highest recall and lowest loss, while Hebbian updates achieve comparable performance with slightly lower plastic norms.  
Mean neuromodulation $\eta(t)$ remains high for the gradient rule ($8.4{\times}10^{-2}$) and near zero for Hebbian ($4.3{\times}10^{-4}$), indicating persistent versus event-gated adaptation.

\subsection{Cue–Reward Association}
\textbf{Task.} Learn five random cue–reward pairs presented sequentially.  
\textbf{Metric.} Validation and “query” loss (error during recall phase).

\begin{table}[t]
\caption{Cue–reward association. Lower loss is better.}
\label{tab:cue}
\begin{center}
\begin{tabular}{lcc}
\multicolumn{1}{c}{\bf RULE} & \multicolumn{1}{c}{\bf VALIDATION LOSS} & \multicolumn{1}{c}{\bf QUERY LOSS}
\\ \hline \\
Gradient & 0.035 $\pm$ 0.008 & 0.064 $\pm$ 0.020 \\
Hebbian & 0.037 $\pm$ 0.014 & 0.065 $\pm$ 0.036 \\
None & {\bf 0.027 $\pm$ 0.010} & {\bf 0.053 $\pm$ 0.020} \\
\end{tabular}
\end{center}
\end{table}

\noindent
\textbf{Interpretation.}
The non-plastic baseline slightly outperforms both plastic variants, suggesting that the low episodic entropy (five cue–reward pairs, $\approx11$ bits) fits within the static weights.  
Gradient-based plasticity maintains higher $\eta(t)$ ($9.3{\times}10^{-2}$) than Hebbian ($2.3{\times}10^{-4}$), but this persistent gating provides no benefit under low-information conditions.

\subsection{One-Shot Image Classification}
\textbf{Task.} 5-way, 1-shot sequential classification on CIFAR-FS and Omniglot.  
\textbf{Episode formatting.} Each episode consists of a support phase (\(N\!\times\!K\) labeled examples) followed by a query phase (\(N\!\times\!Q\) unlabeled examples). For each support image, we encode it into an embedding vector and feed a step vector formed by concatenating \([\text{embedding};\; \text{one-hot(label)};\; 0]\). For each query image, we feed \([\text{embedding};\; \mathbf{0};\; 1]\), where the final scalar is a query flag. The Transformer emits class logits at every step; the task loss (cross-entropy) is applied \emph{only} on query steps, while support steps contribute only via plastic updates. Plasticity (Hebbian or gradient-based) updates the fast weights at every step; the outer-loop gradient is applied once per episode.
\textbf{Setup.} A Conv-4 encoder provides embeddings consumed sequentially by the Transformer.  
\textbf{Metric.} Validation accuracy (higher is better).

\begin{table}[t]
\caption{5-way, 1-shot classification accuracy ($\pm$ std).}
\label{tab:cifarfs}
\begin{center}
\begin{tabular}{lcc}
\multicolumn{1}{c}{\bf RULE} & \multicolumn{1}{c}{\bf CIFAR-FS} & \multicolumn{1}{c}{\bf OMNIGLOT}
\\ \hline \\
Gradient & 0.274 $\pm$ 0.038 & 0.201 $\pm$ 0.012 \\
Hebbian & {\bf 0.319 $\pm$ 0.017} & {\bf 0.237 $\pm$ 0.028} \\
None & 0.289 $\pm$ 0.028 & 0.192 $\pm$ 0.020 \\
\end{tabular}
\end{center}
\end{table}

\noindent
\textbf{Interpretation.}
Hebbian plasticity achieves the best accuracy on both datasets (CIFAR-FS: +4.6 pp; Omniglot: +3.6 pp over gradient).  
Gradient updates tend to maintain large, persistent $\eta(t)$ values ($0.116$ / $0.061$) that degrade accuracy by overfitting to sparse supervision, while Hebbian plasticity fires only around labelled supports.  
This aligns with the hypothesis that local, associative updates outperform dense gradient surrogates in sparse, class-conditional regimes.  
For reference, plastic RNNs from \citet{duan2023hebbian} achieve 55.5 ± 1.0 

\subsection{Few-Shot Regression}
\textbf{Task.} Infer a mapping $f:\mathbb{R}^d\!\to\!\mathbb{R}$ from $K{=}10$ support and $K{=}10$ query pairs.  
\textbf{Metric.} Query mean-squared error (MSE).

\begin{table}[t]
\caption{Few-shot regression ($K{=}10$). Lower is better.}
\label{tab:regression}
\begin{center}
\begin{tabular}{lcc}
\multicolumn{1}{c}{\bf RULE} & \multicolumn{1}{c}{\bf VAL. MSE} & \multicolumn{1}{c}{\bf QUERY MSE}
\\ \hline \\
Gradient & 0.823 $\pm$ 0.019 & 1.589 $\pm$ 0.038 \\
Hebbian & {\bf 0.798 $\pm$ 0.036} & {\bf 1.546 $\pm$ 0.064} \\
None & 1.031 $\pm$ 0.110 & 1.997 $\pm$ 0.266 \\
\end{tabular}
\end{center}
\end{table}

\noindent
\textbf{Interpretation.}
Both plastic rules outperform the baseline, with Hebbian achieving the lowest error despite smaller $\eta(t)$ values ($2.8{\times}10^{-4}$ vs.\ $0.115$ for gradient).  
This supports the view that strong, persistent modulation is not necessarily beneficial in low-signal regimes.

\subsection{Comparison with Plastic RNNs}
To contextualize Transformer results, Table~\ref{tab:duan-comparison} compares performance against plastic RNNs from \citet{duan2023hebbian} on identical benchmarks.

\begin{table}[t]
\caption{Cross-architecture comparison with plastic RNNs \citep{duan2023hebbian}.}
\label{tab:duan-comparison}
\begin{center}
\begin{tabular}{llcc}
\multicolumn{1}{c}{\bf TASK} & \multicolumn{1}{c}{\bf RULE} & \multicolumn{1}{c}{\bf TRANSFORMER (OURS)} & \multicolumn{1}{c}{\bf RNN}
\\ \hline \\
CIFAR-FS acc. & Gradient & 27.4 $\pm$ 3.8\% & 51.2 $\pm$ 2.6\% \\
CIFAR-FS acc. & Hebbian & {\bf 31.9 $\pm$ 1.7\%} & {\bf 55.5 $\pm$ 1.0\%} \\
CIFAR-FS acc. & None & 28.9 $\pm$ 2.8\% & 39.9 $\pm$ 0.8\% \\
Regression MSE & Gradient & 1.589 $\pm$ 0.038 & 0.301 $\pm$ 0.001 \\
Regression MSE & Hebbian & {\bf 1.546 $\pm$ 0.064} & {\bf 0.378 $\pm$ 0.059} \\
Regression MSE & None & 1.997 $\pm$ 0.266 & 0.605 $\pm$ 0.002 \\
\end{tabular}
\end{center}
\end{table}

\noindent
Plastic RNNs retain a headline advantage, yet the relative rule ordering remains consistent, confirming cross-architecture generality of the observed trends.

\subsection{Ablations and Diagnostics}

Targeted ablations clarify dependence on each mechanism.  
Removing the internal gradient target (\texttt{aux\_dim=0}) in the copying task increases loss to $0.362 \pm 0.013$ and lowers recall to $0.735 \pm 0.001$, halving the gradient model’s advantage.  
Freezing Hebbian neuromodulation ($\eta_0=0$) on CIFAR-FS reduces accuracy to $0.296 \pm 0.032$, matching the non-plastic baseline ($0.289 \pm 0.028$).  
Thus, gradient-based plasticity relies on its self-generated targets, whereas Hebbian updates depend critically on adaptive gating.

\begin{figure}[t]
\centering
\includegraphics[width=0.85\textwidth]{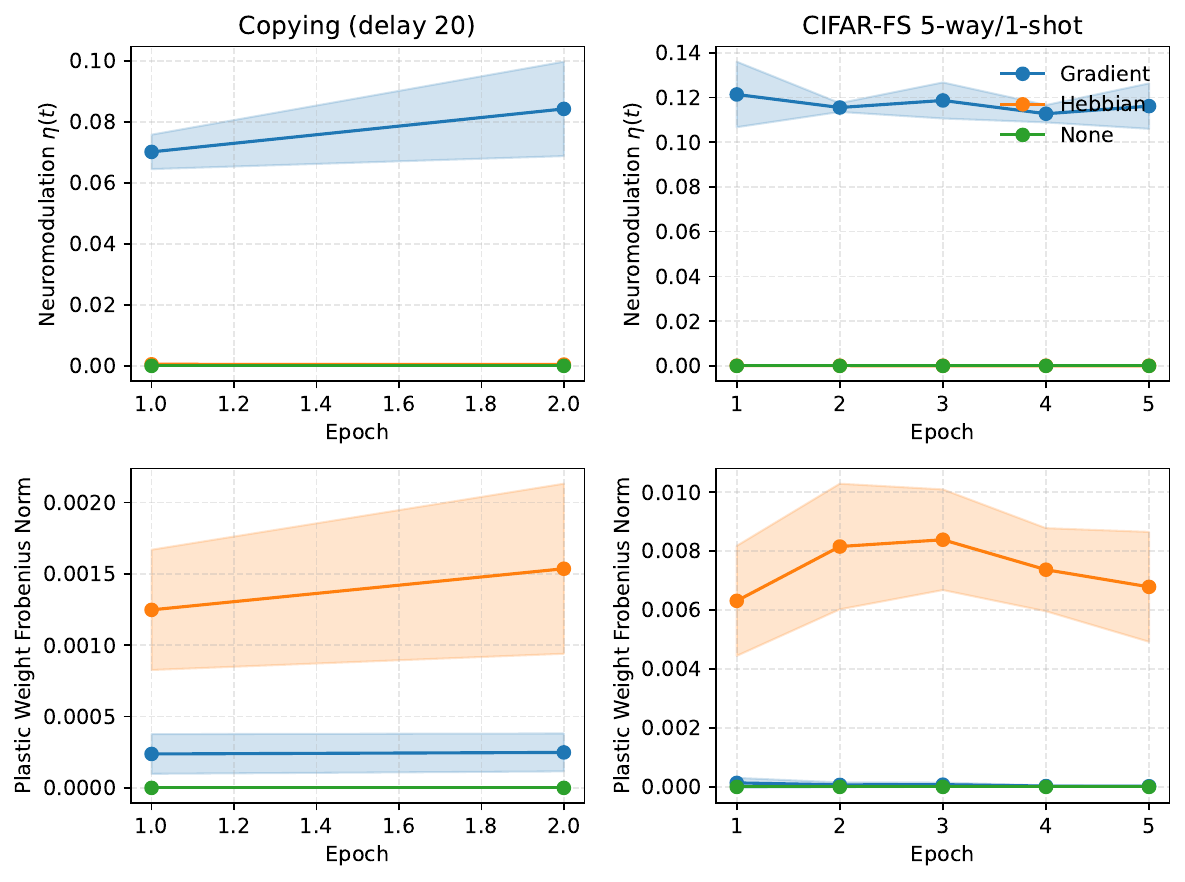}
\caption{Neuromodulation and plastic-weight norms per epoch for copying (left) and CIFAR-FS (right). Solid curves show the mean over three seeds; shaded regions denote one standard deviation. Gradient plasticity sustains high $\eta(t)$ throughout the episode, whereas Hebbian updates fire in short bursts around support tokens.}
\label{fig:mechanistic}
\end{figure}

Mechanistic traces (Fig.~\ref{fig:mechanistic}) show that gradient plasticity maintains high $\eta(t)$ ($8.4{\times}10^{-2}$ on copying, $1.16{\times}10^{-1}$ on CIFAR-FS) with small plastic norms ($2.3{\times}10^{-5}$), while Hebbian updates exhibit sparse bursts ($\eta(t){\sim}10^{-4}$) but larger weight norms ($6.8{\times}10^{-3}$).  
The non-plastic baseline yields zero modulation and serves as a control.

\subsection{Task-Dependent Behaviour and Depth Stress Test}

Table~\ref{tab:taxonomy} summarises how each rule performs under varying supervision density and episodic entropy.

\begin{table}[t]
\caption{Empirical taxonomy of plasticity performance.}
\label{tab:taxonomy}
\begin{center}
\begin{tabular}{lcccp{5cm}}
\multicolumn{1}{c}{\bf TASK} & 
\multicolumn{1}{c}{\bf DENSITY} & 
\multicolumn{1}{c}{\bf EPISODIC ENTROPY} & 
\multicolumn{1}{c}{\bf WINNER} & 
\multicolumn{1}{c}{\bf EXPLANATION}
\\ \hline \\
Copying & Dense & High & Gradient & Each token provides a target; long-range memory required. \\
Cue–reward & Dense & Low & None & Static weights suffice for small associative sets. \\
Classification & Sparse & High & Hebbian & Sparse supervision; local outer-product writes aid recall. \\
Regression & Sparse & High & Hebbian & Support examples encode latent functions needing fast storage. \\
\end{tabular}
\end{center}
\end{table}

Extending the copying task to 8-layer models exposes stability limits.  
Gradient-plastic Transformers diverge after $\sim$3000 steps (plastic norms $>10^2$; recall below baseline), with the deepest layers showing the largest drift.  
Hebbian plasticity remains stable but saturates in performance (recall $0.729 \pm 0.015$).  
A practical regime therefore lies around 4 layers: deeper gradient-plastic stacks require additional regularization (e.g., gradient clipping or frozen upper layers) to prevent instability.

\section{Discussion}

Our experiments demonstrate that explicit, biologically inspired plasticity can be integrated into standard decoder-only Transformers with minimal architectural modification. Across tasks, the results support a consistent functional taxonomy linking supervision density and episodic complexity to the most effective plasticity rule.

\paragraph{Summary of empirical trends.}
\begin{itemize}
    \item \textbf{Gradient-based plasticity} excels on densely supervised, delay-heavy sequences such as the copying task. When every token provides a regression signal, the gradient rule effectively propagates credit across long horizons and yields the highest recall even in compact 4-layer models.
    \item \textbf{Hebbian plasticity} dominates in exemplar-driven or sparsely supervised settings. On few-shot classification and regression, outer-product updates embed support examples directly into the fast weights, achieving the lowest error while operating with two orders of magnitude smaller neuromodulation.
    \item \textbf{Non-plastic baselines} suffice when episodic entropy is low, as seen in the cue–reward task where static parameters can memorize the small set of associations encountered during meta-training.
    \item \textbf{Mechanistic signals as diagnostics.} Monitoring $\eta(t)$ and fast-weight norms provides a useful lens on model behaviour: persistently high $\eta(t)$ during queries indicates overactive gradient plasticity, while near-zero modulation suggests Hebbian saturation or underuse.
\end{itemize}

\paragraph{Conceptual implications.}
The observed task-dependent behaviour aligns with theoretical accounts of in-context optimization in Transformers \citep{akyurek2023learning, vonoswald2023transformers}. Dense credit assignment regimes favour global, gradient-like update mechanisms, whereas sparse or discrete supervision naturally engages local correlation-based rules. In this sense, the two plasticity forms occupy complementary positions along a continuum from implicit in-context learning to explicit fast-weight adaptation. The results also reinforce that explicit plasticity can coexist stably with self-attention and provides a controllable handle on when and how adaptation occurs within a sequence.

\paragraph{Practical guidelines.}
Gradient plasticity is most effective when feedback is continuous or graded (e.g., sequence prediction, delayed regression). Hebbian plasticity is preferable for sparse, class-conditional supervision where labelled supports must be written into short-term memory. Explicit plasticity is unnecessary when the per-episode information content is low or static fine-tuning already captures the training distribution.

\paragraph{Limitations and future work.}
The gradient-based rule required short training horizons to remain numerically stable; scaling to longer contexts may need truncated inner-loop backpropagation or additional regularization on the plastic buffers. While three-seed sweeps are sufficient to establish consistent rule ordering, larger-scale runs would narrow confidence intervals. Finally, direct comparisons with large pre-trained Transformers performing purely in-context prompting remain an open direction: such models represent a different training regime from the meta-learned setting studied here, and bridging the two could clarify how emergent and explicit adaptation interact in large-scale systems.

\section{Code and Compute}

\paragraph{Code.}
All experiments were implemented in \texttt{Python~3.10} using \texttt{PyTorch~2.1.1} and the Hydra configuration framework.  
Each benchmark (copying, cue–reward, few-shot regression, CIFAR-FS, and Omniglot classification) is defined as an independent configuration module, with runs launched via unified command-line wrappers across three random seeds (3000–3002).  
Raw training logs, per-epoch diagnostics, and aggregated metrics are automatically stored as JSON files, enabling full reproducibility.  
All figures and tables were generated by a single plotting script that consumes these logs and outputs publication-ready assets.  
The complete codebase, including configurations and pretrained checkpoints, is available at the project repository:  
\url{https://github.com/sidcraftscode/hebbian-transformer/}.

\paragraph{Compute.}
All runs were executed on a shared Linux server with a single NVIDIA A100 GPU (40\,GB VRAM) and 64\,GB host RAM.  
Typical three-seed sweeps required:
\begin{itemize}
    \item \textbf{Copying / Cue–Reward:} $\approx$6\,min per three-seed bundle (peak 7.6\,GB VRAM).
    \item \textbf{Few-Shot Regression:} $\approx$12\,min (peak 9.1\,GB VRAM).
    \item \textbf{CIFAR-FS / Omniglot Classification:} up to 45\,min for three seeds with the Conv-4 encoder (peak 18\,GB VRAM).
\end{itemize}
Mechanistic diagnostics (neuromodulation traces and plastic-weight norms) are recorded on-the-fly during training, requiring no separate reruns.  
Across all experiments, the cumulative compute requirement was approximately 25\,GPU-hours on a single NVIDIA~A100, including all three-seed repetitions and post-processing.
This moderate budget reflects the efficiency of the experimental design and makes the full study reproducible on a single-GPU workstation.

\section{Conclusion}

This work demonstrates that explicit synaptic plasticity can be seamlessly integrated into autoregressive Transformers, yielding models that adapt within a single sequence rather than across gradient updates. By equipping feed-forward layers with fast-weight components updated through either neuromodulated Hebbian rules or gradient-based plasticity, we show that self-attention architectures can emulate the rapid, context-dependent learning observed in biological systems.

Empirically, the two mechanisms occupy complementary functional regimes. Gradient-based plasticity thrives when dense, continuous feedback supports distributed credit assignment, while Hebbian updates dominate under sparse, event-driven supervision such as few-shot classification or regression. These results outline a clear taxonomy of when and how explicit adaptation improves in-context learning, bridging emergent behavior in large Transformers with the mechanistic interpretability of biologically grounded models.

Beyond task performance, the proposed framework introduces a controllable axis of adaptation that can be probed, gated, and regularized. This opens opportunities for scaling explicit plasticity to large pre-trained models, combining it with recurrent or memory-augmented architectures, and using it to study alignment between learned and biological learning dynamics. Future directions include extending plasticity to attention weights themselves, investigating stability at greater depth or context lengths, and exploring hybrid systems where implicit and explicit adaptation coexist within a unified learning architecture.

Ultimately, the findings suggest that synaptic plasticity—long viewed as a biological metaphor—can serve as a practical design principle for building more adaptive, interpretable, and energy-efficient Transformer systems.

\bibliography{tmlr}
\bibliographystyle{tmlr}

\appendix
\section{Implementation Details}

\subsection{Model and Encoder Hyperparameters}
All experiments use the two-layer decoder-only Transformer defined in Section~\ref{sec:method}.  
Copying, cue--reward, and regression tasks operate directly on low-dimensional inputs, so we adopt a compact configuration ($d_{\text{model}}=128$, $d_{\text{ff}}=256$, four heads).  
Image classification employs the Conv-4 encoder in \texttt{src/models/conv\_encoder.py}---four Conv-BN-ReLU-MaxPool blocks with 64 channels followed by a linear projection to 256 dimensions---and widens the Transformer to $d_{\text{model}}=256$ and $d_{\text{ff}}=512$.  
The auxiliary head is fixed at dimension~4 so that gradient-based plasticity can synthesise an internal loss. Dropout is $0.1$ throughout.

\begin{table}[h]
\caption{Transformer configuration by task.}
\label{tab:appendix-model-config}
\begin{center}
\begin{tabular}{lccccc}
\multicolumn{1}{c}{\bf TASK} & 
\multicolumn{1}{c}{\boldmath{$d_{\text{model}}$}} & 
\multicolumn{1}{c}{\boldmath{$d_{\text{ff}}$}} & 
\multicolumn{1}{c}{\bf HEADS} & 
\multicolumn{1}{c}{\bf LAYERS} & 
\multicolumn{1}{c}{\bf AUX DIM}
\\ \hline \\
Copying / Cue--reward / Regression & 128 & 256 & 4 & 2 & 4 \\
CIFAR-FS / Omniglot classification & 256 & 512 & 4 & 2 & 4 \\
\end{tabular}
\end{center}
\end{table}

\subsection{Optimisation and Training Schedules}
We train with AdamW (PyTorch 2.1.1), learning rate $10^{-3}$, weight decay $5\times10^{-4}$ for classification and $10^{-4}$ elsewhere, and gradient clipping at 5.0.  
All plastic models use $\eta_0=0.2$ and $\text{max\_norm}=1.0$.  
Each configuration is run with three seeds (3000--3002); validation occurs at the end of every epoch for sequence tasks and after every 20 (regression) or 50 (classification) episodes.  
Table~\ref{tab:appendix-training} summarises the schedule.

\begin{table}[h]
\caption{Training schedule and episode counts. Batch size is one sequence for the sequence tasks and the full support/query set for classification episodes.}
\label{tab:appendix-training}
\begin{center}
\begin{tabular}{p{3.2cm}p{4.2cm}p{1.2cm}p{3cm}}
\multicolumn{1}{c}{\bf TASK} &
\multicolumn{1}{c}{\bf TRIALS / EPISODES PER EPOCH} &
\multicolumn{1}{c}{\bf EPOCHS} &
\multicolumn{1}{c}{\bf DATASET SIZE}
\\ \hline \\
Copying ($n{=}5$, delay 20) & 50 sequences & 2 & 50 \\
Cue--reward (8 pairs) & 100 sequences & 4 & 100 \\
Few-shot regression ($K{=}10$) & 150 functions & 5 & 150 \\
CIFAR-FS classification & 80 episodes (5-way, 1-shot, 15 queries) & 5 & 80 train / 50 val \\
Omniglot classification & 80 episodes (5-way, 1-shot, 15 queries) & 5 & 80 train / 50 val \\
\end{tabular}
\end{center}
\end{table}

\subsection{Task Generation}
\paragraph{Copying.} Sequences comprise five discrete symbols drawn from a 10-token vocabulary, followed by a 20-step blank delay and a recall phase flagged by a binary indicator.

\paragraph{Cue--reward.} Each 20-step episode samples eight cues from $[0,1]^{20}$ together with rewards in $[0,1]$. Rewards are revealed during the first 10 steps and must be recalled during the remaining steps.

\paragraph{Few-shot regression.} For every meta-training task we sample a random affine map $\mathbb{R}^3\rightarrow\mathbb{R}$ with weight and bias scale~1.0, observe 10 noisy support pairs, and evaluate on 10 held-out query pairs.

\paragraph{Classification.} CIFAR-FS and Omniglot episodes use 5-way, 1-shot support sets with 15 query images per class. Support embeddings (from the Conv-4 encoder) and one-hot labels are interleaved before the queries in the Transformer input stream.

\section{Code Availability}
\label{sec:code-availability}

The code accompanying this paper is available at:\\[1mm]
\url{https://github.com/sidcraftscode/hebbian-transformer/}

\label{sec:appendix-figures}
\clearpage
\FloatBarrier
\section{Supplementary Figures}
\label{sec:appendix-figures}

\begin{figure}[h]
\centering
\includegraphics[width=0.92\textwidth]{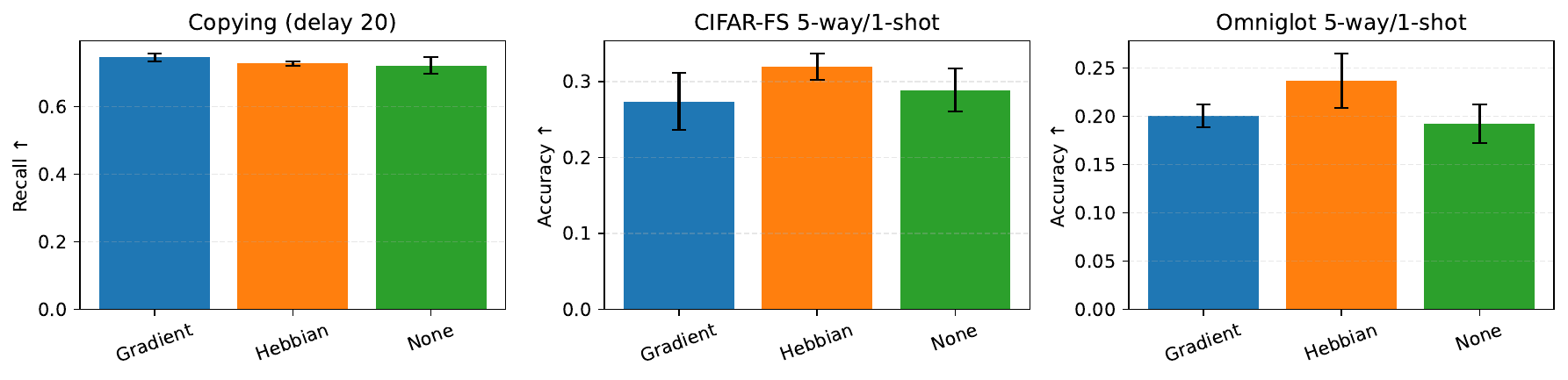}\\[3mm]
\includegraphics[width=0.92\textwidth]{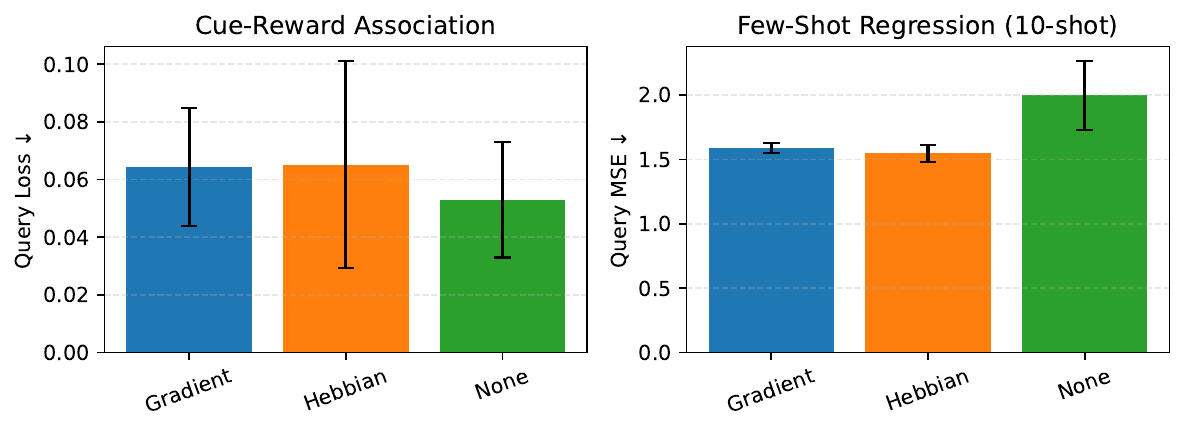}
\caption{Aggregate validation metrics across tasks. \textbf{Top:} Accuracy-style objectives (higher is better) for copying recall and the two classification datasets. \textbf{Bottom:} Loss-style objectives (lower is better) for cue–reward query loss and few-shot regression mean-squared error. Bars show the mean across three seeds; whiskers denote one standard deviation.}
\label{fig:performance-summary}
\end{figure}

\end{document}